4# The possibility of making $138,000 from shredded banknote pieces using computer vision

Chung T. Kong, Member IEEE*Abstract*—Every country must dispose of old banknotes. At the Hong Kong Monetary Authority visitor center, visitors can buy a paperweight souvenir full of shredded banknotes. Even though the shredded banknotes are small, by using computer vision, it is possible to reconstruct the whole banknote like a jigsaw puzzle. Each paperweight souvenir costs $100 HKD, and it is claimed to contain shredded banknotes equivalent to 138 complete $1000 HKD banknotes. In theory, $138,000 HKD can be recovered by using computer vision. This paper discusses the technique of collecting shredded banknote pieces and applying a computer vision program.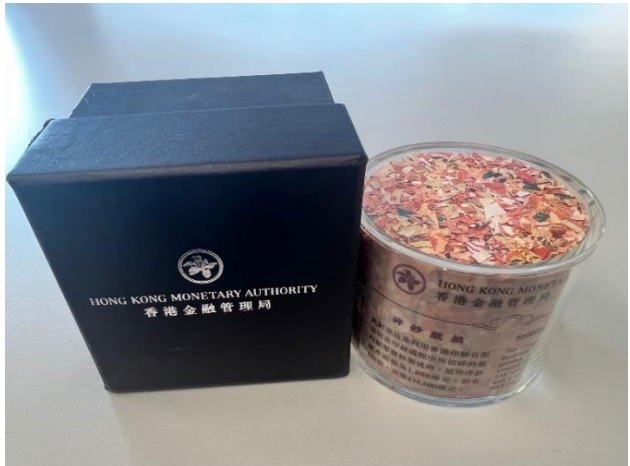

*Fig. 1: Shredded-banknote paperweight souvenir from the Hong Kong Monetary Authority.*

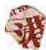

*Fig. 2a: A shredded banknote piece.*

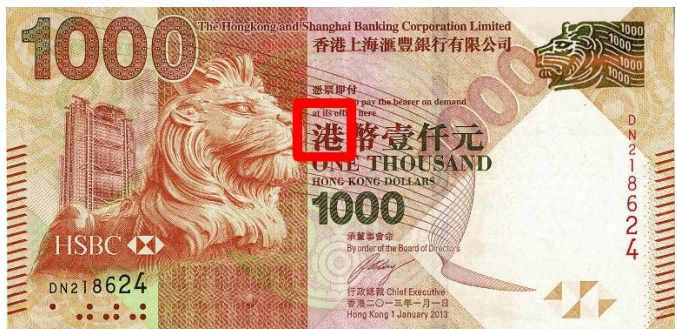

*Fig. 2b: The shredded banknote piece from panel (a) located on a banknote with a red box.*

## I. INTRODUCTION

Every country must dispose of old banknotes. At the Hong Kong Monetary Authority visitor center, visitors can buy a paperweight souvenir full of shredded banknotes. Even though the pieces are small, by using computer vision, it is possible to reconstruct the whole banknote like a jigsaw puzzle.

In Hong Kong, only three banks may print bank notes. They are HSBC, Bank of China, and Standard Charter Bank. In the Hong Kong Monetary Authority visitor center, each visitor can purchase a maximum of two paperweight souvenirs.

The paperweight souvenir is filled with shredded 2018 banknotes from these three banks. Information about the paperweight is publicly available at the Hong Kong Monetary Authority visitor center[1].

The basic concept of this research was to use an entire banknote as the ground truth. For each shredded banknote piece, a computer program mapped the piece to the ground truth. A red rectangle on the banknote served as the output to identify the location and orientation of the shredded piece.

## II. BACKGROUND

Many countries, such as the United States[2] and Saudi Arabia[3], sell shredded retired banknotes as souvenirs. In Hong Kong, the Hong Kong Monetary Authority visitor center sells a paperweight souvenir full of shredded banknotes. In theory, even if the shredded banknote pieces are small, it is possible to recreate an entire banknote from these shredded banknote pieces. A proof of concept using computer vision is presented in this paper so that the banking industry can be aware of this loophole in creating money.

According to The Association of Commercial Banknote Issuers [4], "at least half of the banknote and visible serial number." Obtaining a serial number is a challenge in regenerating a banknote from shredded pieces because the paperweight may not come from the same banknote. In addition, each banknote has a unique serial number. Thus, exchanging shredded banknote pieces for real money at a bank may not be possible. Nevertheless, this paper focuses on converting the shredded banknotes back to a complete banknote.

The shredded banknote pieces were enclosed in a cylinder of proxy glass in a cylindrical container. Proxy glass melts at approximately 200°C; a household hair dryer would be sufficient to soften this glass. Fig. 3 shows the lid being removed using a heat gun.

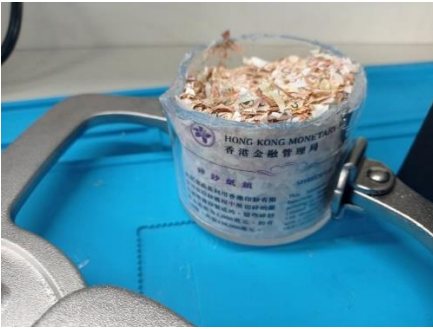

*Fig. 3: The paperweight lid was opened using a heat gun.*

The shredded banknotes were then collected. Surprisingly, three paperweight cylinders were opened, and two of them had stones in them (Figs. 4 and 5)!

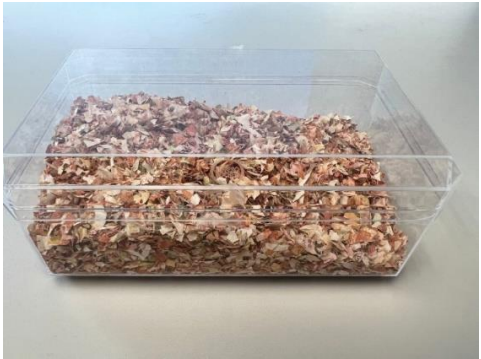

*Fig. 4: The shredded banknotes were collected.*

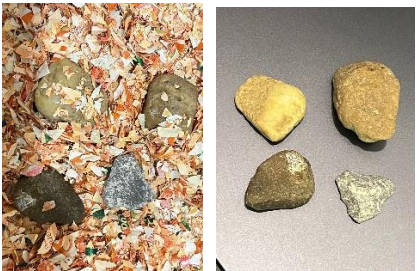

*Fig. 5: Stones inside one of the containers.*

Before employing any computer vision, I attempted to put the pieces together manually. In theory, if time is not a limiting factor, a human could put these shredded pieces back together. I spent one day putting some pieces together (Fig. 6).

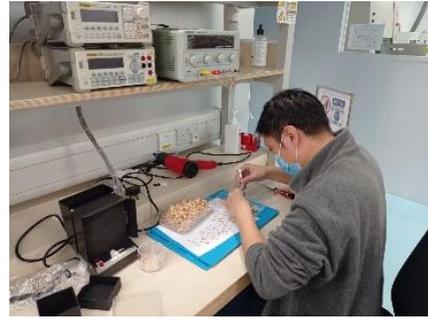

*Fig. 6: The author working on the shredded banknotes.*

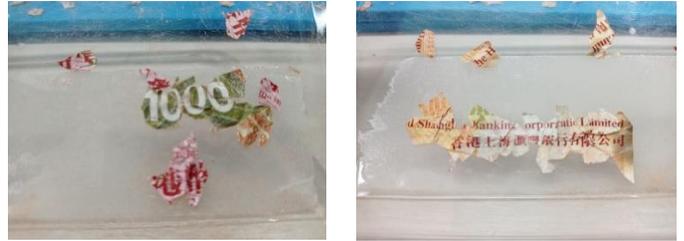

*Fig. 7: The left panel shows the word "$1000," and the right panel shows the HSBC name written in Chinese.*

The serial number is a critical element of the banknote; therefore, it is important to search for pieces with a serial number. Luckily, in the pile of shredded banknote pieces, there were many pieces containing serial numbers. However, there was no evidence that these pieces had come from the same banknote.

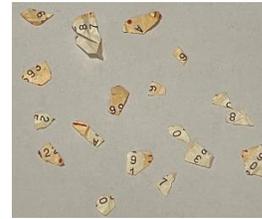

*Fig. 8: A collection of shredded banknotes with serial numbers.*

Several methods have been described to solve jigsaw puzzles using computer vision [5-7]. The difference between the goal of this project and solving a jigsaw puzzle was the irregular shape of the shredded banknote pieces and the quality of the pieces. Several data preparation steps needed to be processed before using computer vision to increase the chance of having a match on the banknotes.

### III. SCOPE OF WORK

The following description of the scope of work has been divided into two main sections: machine learning and mapping. The purpose of machine learning was to distinguish shredded banknote pieces containing a serial number from the rest of the shredded pieces. The mapping section consisted of three smaller parts: ground truth preparation (part A), shredded banknote rotation and cropping (part B), and application of the computer vision program (part C).

**Machine Learning:**

Because the serial number is unique on every banknote, it was necessary to identify any pieces with serial numbers. The machine learning training and inference were based on Dusty-nv inference[8] using the Nvidia Nano Development Board[9]. The classification contained two classes, namely regular class and serial number class, labeled as R and S, respectively.

Only 100 images of each class were required to train the machine learning model using "googlenet" with 30 epochs. Fig. 9 presents some example results.

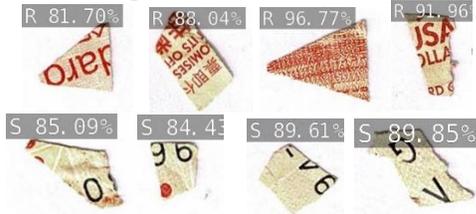

*Fig. 9: The top four photographs show the inference correctly identified as regular class. The bottom four photographs show the inference correctly identified as serial number class.*

After the machine learning step, the pieces with serial numbers were extracted. There were only two possible locations on each banknote for pieces with a serial number; therefore, placing the pieces manually was not difficult. The regular pieces were fed into the mapping section; Fig. 10 presents a block diagram of this system.

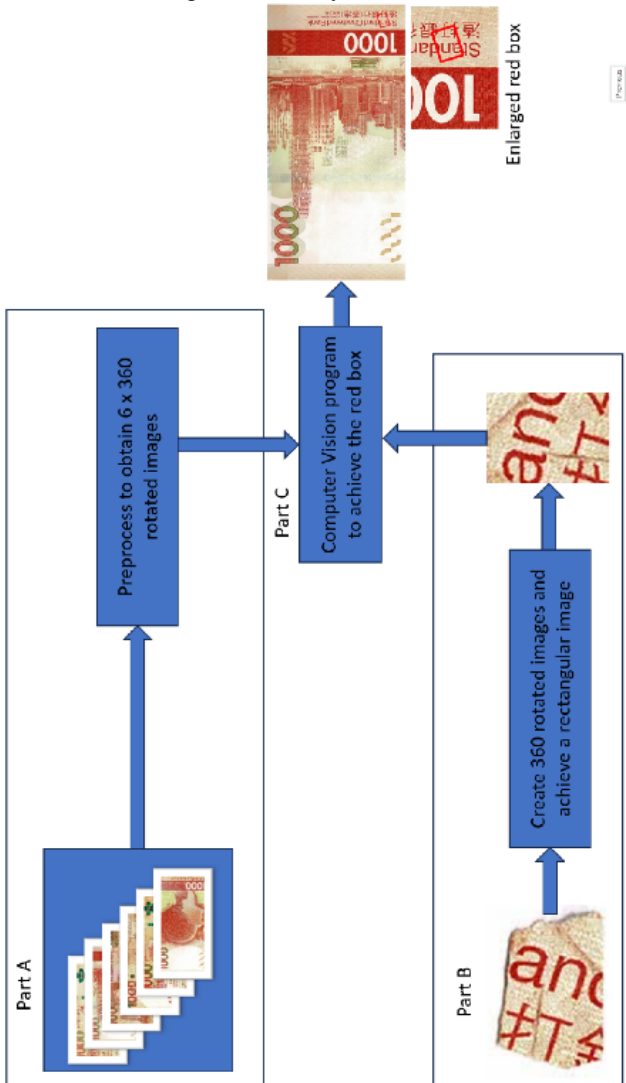

*Fig. 10: Block diagram of the mapping system.*

*Part A is ground-truth preprocessing. Part B is shredded banknote rotation and cropping. Part C is use of the computer vision program.*

**Mapping:**

In part A, three banknotes from three different banks were scanned on the computer. Six images formed the ground truth (Fig. 11).

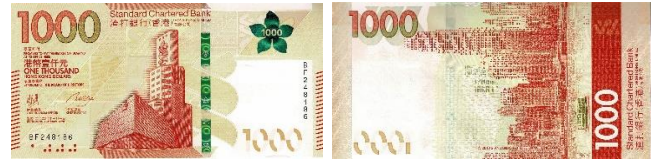

*Fig. 11a: Standard Charter banknote.*

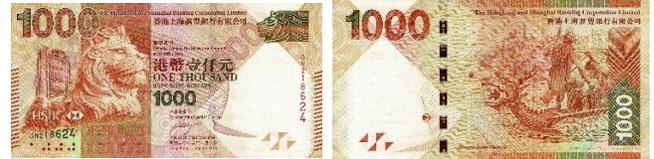

*Fig. 11b: HSBC banknote.*

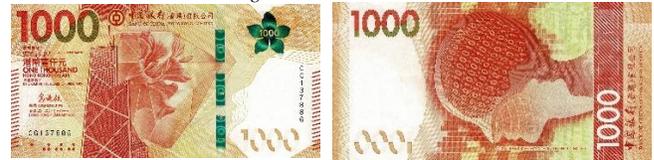

*Fig. 11c: Bank of China banknote.*

In part B, each shredded banknote piece was scanned and then cloned into 360 images with full rotation. From all the rotated images, the largest rectangular image with no white edges was saved as a template (Fig. 12).

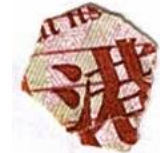

*Fig. 12a: Shredded banknote piece (not to scale).*

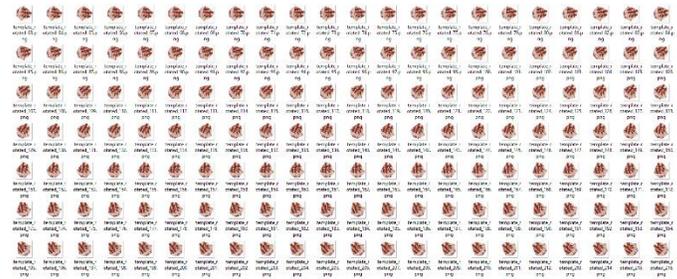

*Fig. 12b: Rotated shredded banknote piece.*

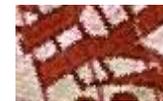

*Fig. 12c: Chopped banknote template (not to scale).*

Because the shredded banknote pieces may not have been in the correct orientation, all six ground-truth banknote images were rotated 360 degrees. This process produced $6 \times 360 = 2160$ ground truth images.

Using the program, the rectangular banknote template was mapped onto the 2160 ground truth images using the OpenCV template matching function.

The best result was labeled on the ground-truth banknote with a red box. Below is the result of an example. The matching value is the similarity; *x* and *y* are the coordinates and rotation of the ground truth to match the template:

*match_value = 0.7967000603675842*
*x = 1100*
*y = 1386*
*rotation = 121*

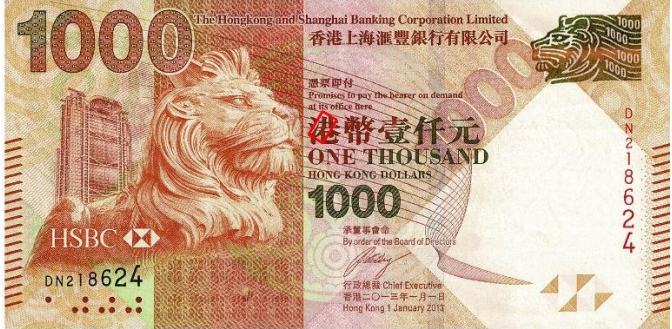
Fig. 13: The best shredded banknote piece position on the HSBC banknote.

## IV. RESULTS

Testing involved 28 shredded banknote pieces. Fig. 14 presents the collection of the pieces.

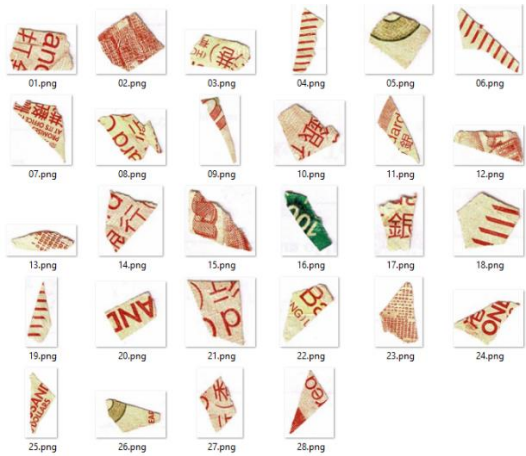
Fig. 14: The collection of shredded banknote pieces.

In the results shown in Fig. 15, the red box shows the position and orientation of each shredded banknote piece on the banknote. In each panel of the figure, the far-right image indicates the shredded banknote piece, the middle image shows the cropped shredded banknote, and the far-left image presents the output location and orientation on the ground truth. (Note: The images in each panel are not to scale.)

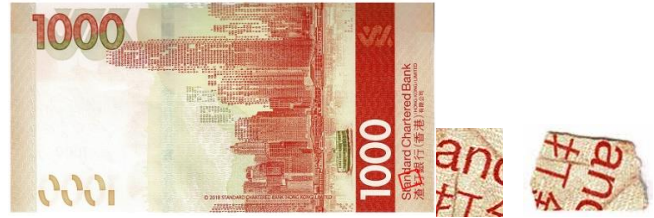
Fig. 15(01): Shredded banknote mapping.

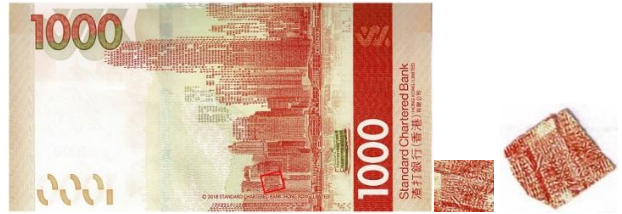
Fig. 15(02): Shredded banknote mapping.

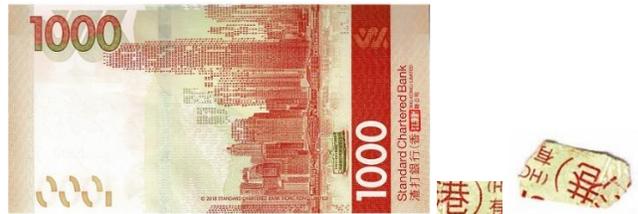
Fig. 15(03): Shredded banknote mapping.

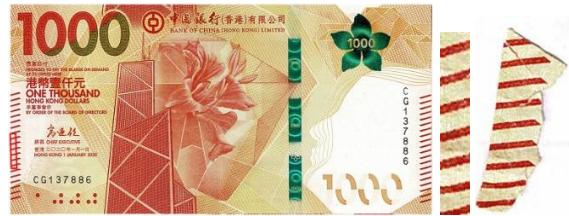
Fig. 15(04): Shredded banknote mapping.

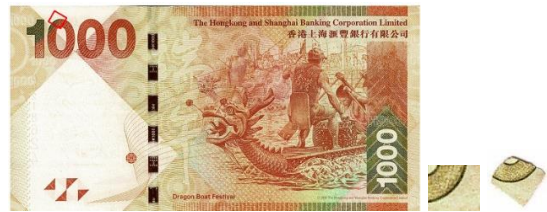
Fig. 15(05): Shredded banknote mapping.

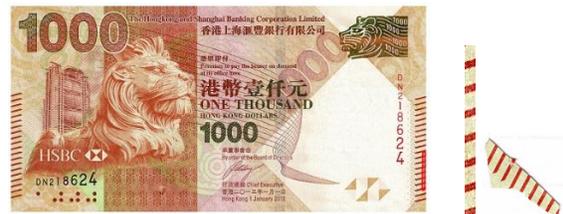
Fig. 15(06): Shredded banknote mapping.

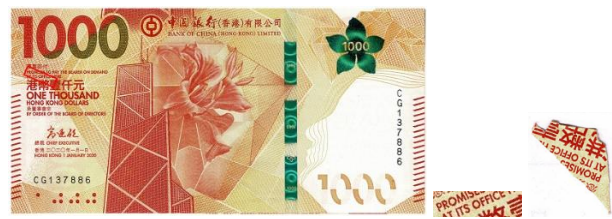

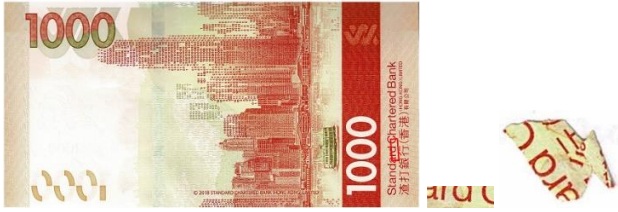

*Fig. 15(07): Shredded banknote mapping.*

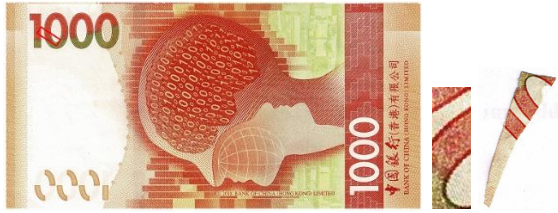

*Fig. 15(08): Shredded banknote mapping.*

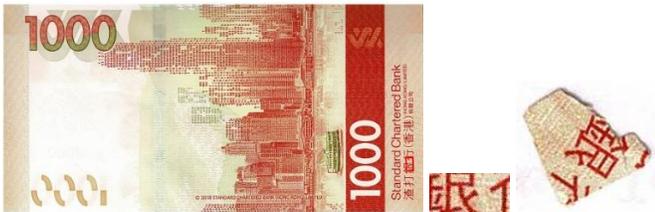

*Fig. 15(09): Shredded banknote mapping.*

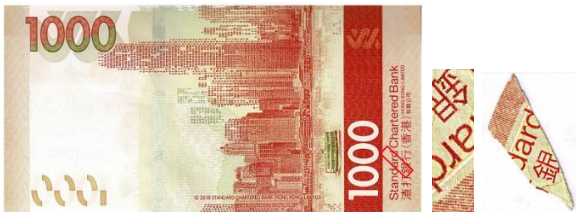

*Fig. 15(10): Shredded banknote mapping.*

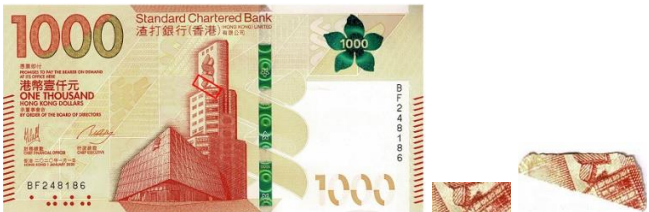

*Fig. 15(11): Shredded banknote mapping.*

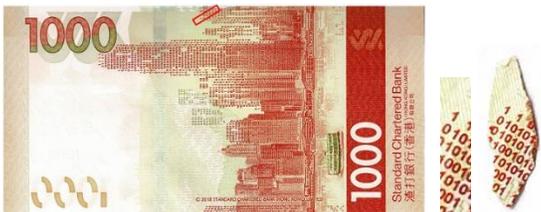

*Fig. 15(12): Shredded banknote mapping.*

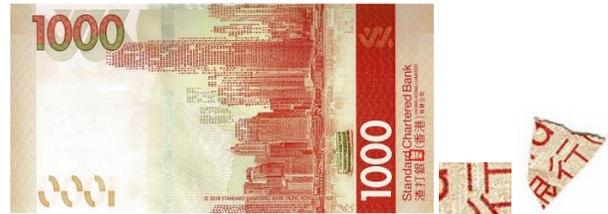

*Fig. 15(13): Shredded banknote mapping.*

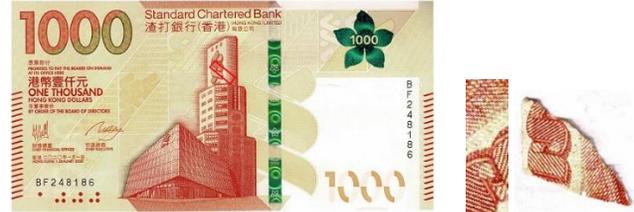

*Fig. 15(14): Shredded banknote mapping.*

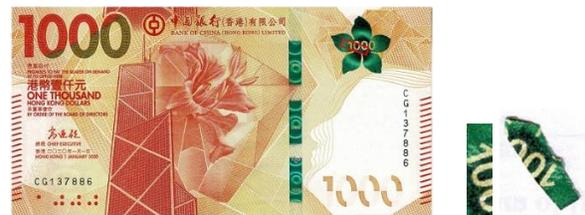

*Fig. 15(15): Shredded banknote mapping.*

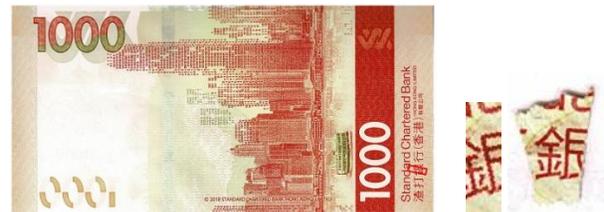

*Fig. 15(16): Shredded banknote mapping.*

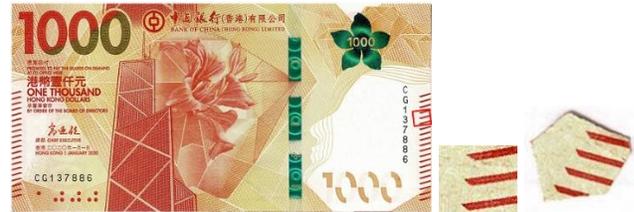

*Fig. 15(17): Shredded banknote mapping.*

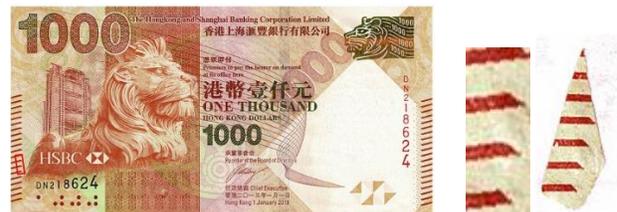

*Fig. 15(18): Shredded banknote mapping.*

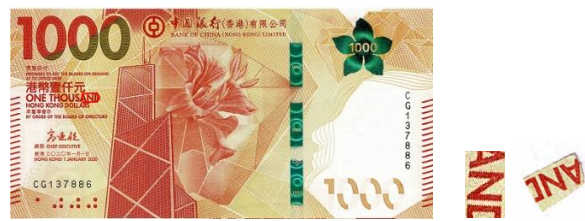

*Fig. 15(19): Shredded banknote mapping.*

*Fig. 15(20): Shredded banknote mapping.*

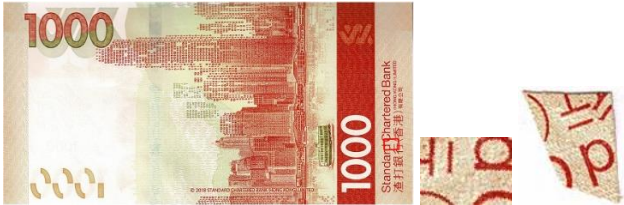

*Fig. 15(21): Shredded banknote mapping.*

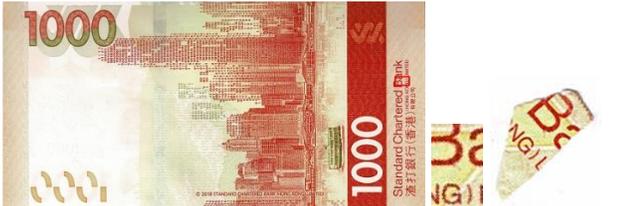

*Fig. 15(22): Shredded banknote mapping.*

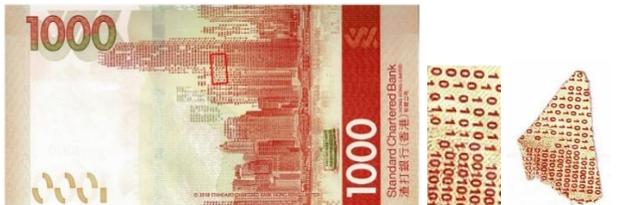

*Fig. 15(23): Shredded banknote mapping.*

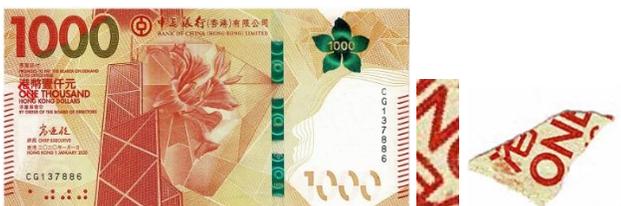

*Fig. 15(24): Shredded banknote mapping.*

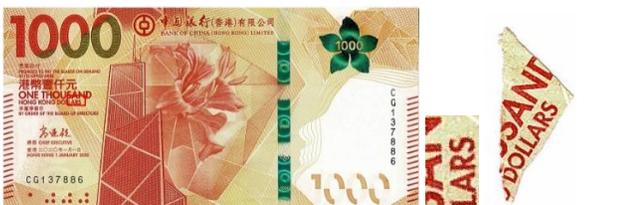

*Fig. 15(25): Shredded banknote mapping.*

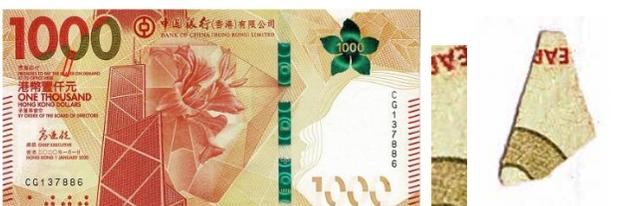

*Fig. 15(26): Shredded banknote mapping.*

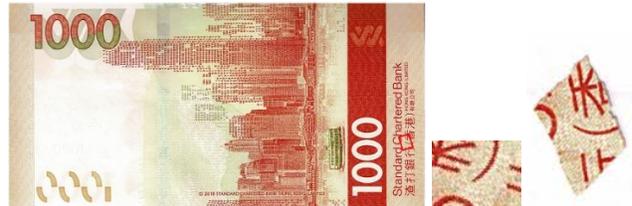

*Fig. 15(27): Shredded banknote mapping.*

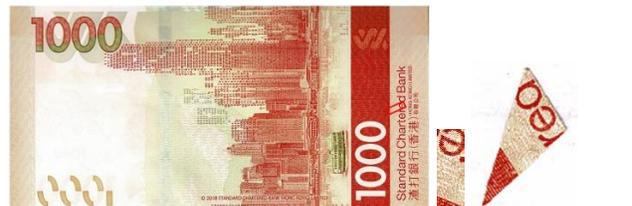

*Fig. 15(28): Shredded banknote mapping.*

## V. ADDITIONAL FINDINGS

Because one of the three cylinders contained some stones, the fourth cylinder was opened with care. The weight was recorded for each step. The label on the cylinder claimed that it contained 138 pieces of equivalent shredded banknotes. By comparing the weight of the shredded banknotes with that of the actual banknote, the number of equivalent shredded banknotes could be calculated.

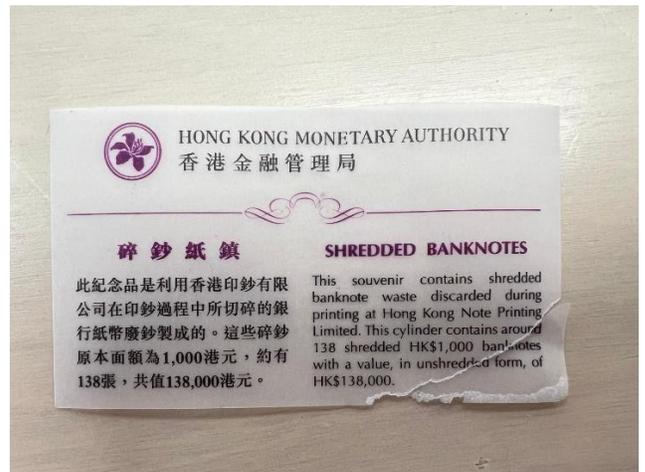

*Fig. 16: The label on the paperweight states, "This cylinder contains around 138 shredded HK$1,000 banknotes with a value, in unshredded form, of HK$138,000."*

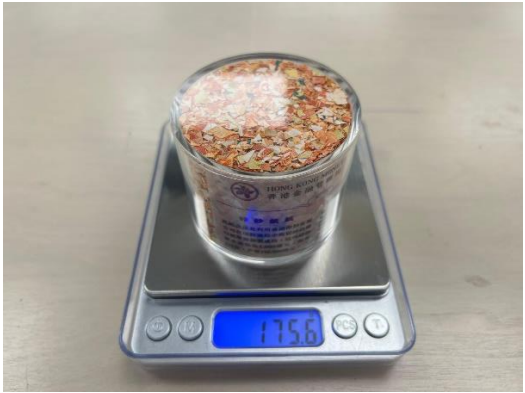
Fig. 17: Unopened paperweight souvenir.

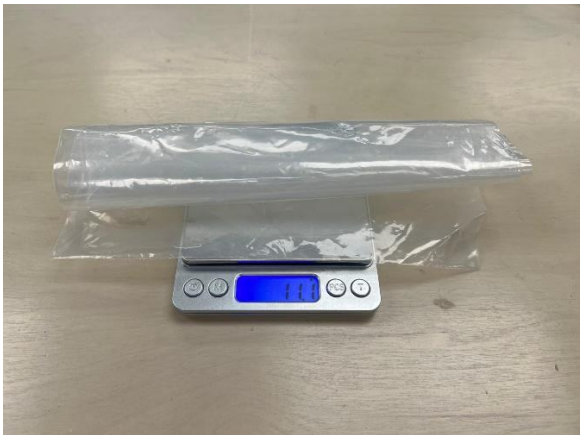
Fig. 18: Empty plastic bag (used to retain the shredded banknote pieces later).

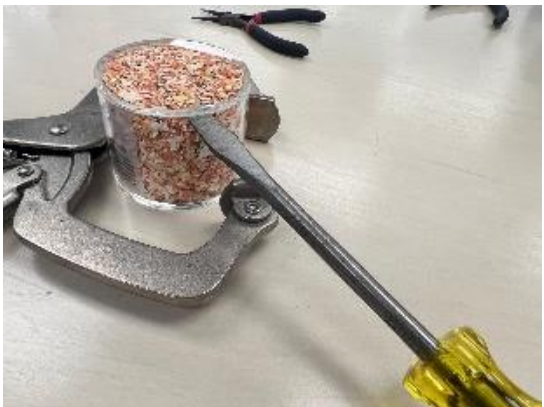
Fig. 19: Using a flathead screwdriver to open the lid.

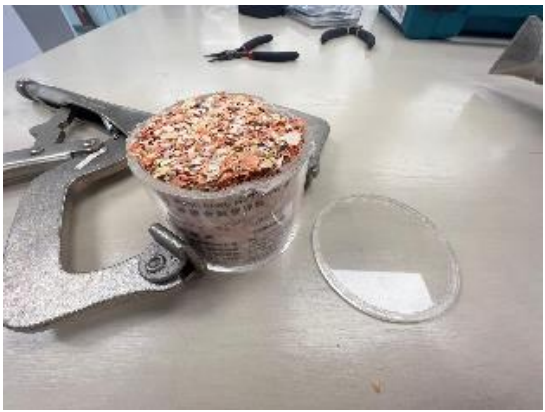

Fig. 20: The container after opening.

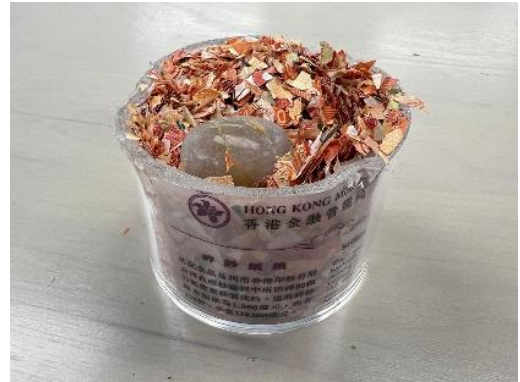
Fig. 21: One of the stones seen in the container.

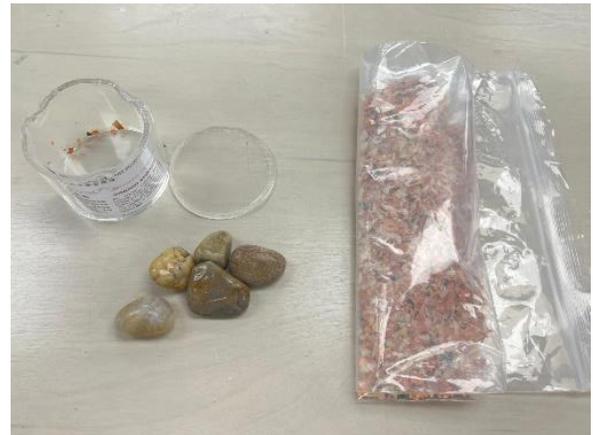
Fig. 22: The complete set. On the upper left side is the empty container. The five stones found in the container are on the lower left side. On the right side is the plastic bag containing all the shredded banknote pieces.

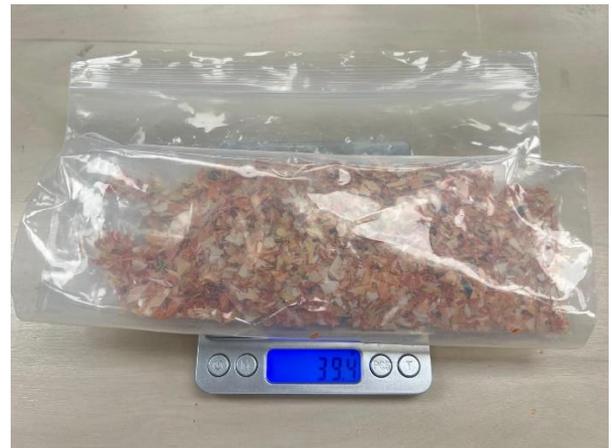
Fig. 23: The weight of the plastic bag containing the shredded banknotes was 39.4 g.

The calculation was as follows:
Weight of the empty plastic bag = 11.1 g
Weight of the plastic bag with shredded banknotes = 39.4 g
Weight of the total shredded banknotes
   = 39.4 g – 11.1 g = only 28.3 g

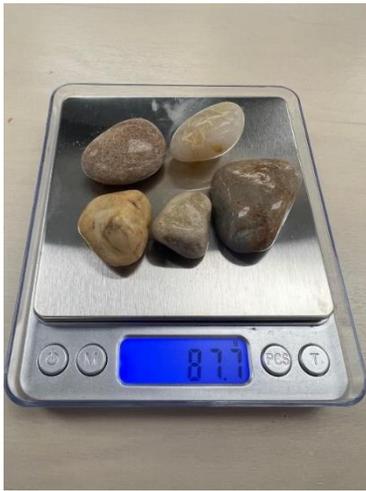

*Fig. 24: The weight of all five stones was 87.7 g.*

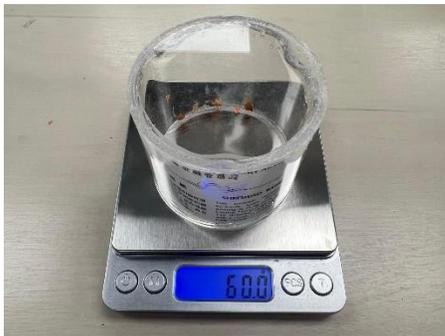

*Fig. 25: The weight of the empty container with the lid was 60.0 g.*

The calculation was as follows:
Weight of the unopened paperweight = 175.6 g
Weight of the five stones = 87.7 g
Weight of the shredded banknotes: 28.3 g
The weight of the empty container = 60 g

To verify the findings:
Weight of unopened paperweight
   = weight of empty paperweight
   + weight of the five stones
   + weight of the shredded banknote pieces
   = 60 + 87.7 + 28.3 = 176 g

This result is approximately the same weight as the unopened paperweight (175.6 g).

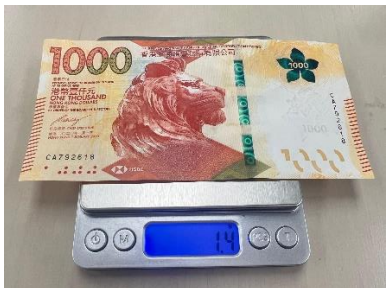

*Fig. 26: For comparison, the weight of one HK$1,000 banknote was 1.4 g.*

A HK$1,000 banknote weighed approximately 1.4 g; therefore, the paperweight contained many fewer than 138 equivalent banknotes.
   = weight of the shredded banknotes/weight of one banknote
   = 28.3/1.4 = approximately 20 equivalent banknotes.

It was surprising that the Hong Kong Monetary Authority had provided only 20 equivalent banknotes rather than 138 equivalent banknotes. The cylinder only contained 20/138 (= 14.5%) of the shredded banknote pieces that the label had claimed.

The Trade Descriptions Ordinance in Hong Kong[10] states the following purpose:

*To prohibit false trade descriptions, false, misleading or incomplete information, false marks and misstatements in respect of goods provided in the course of trade or suppliers of such goods; to confer power to require information or instruction relating to goods to be marked on or to accompany the goods or to be included in advertisements; to restate the law relating to forgery of trade marks; to prohibit certain unfair trade practices; to prohibit false trade descriptions in respect of services supplied by traders; to confer power to require any services to be accompanied by information or instruction relating to the services or an advertisement of any services to contain or refer to information relating to the services; and for purposes connected therewith.*

Two paperweight souvenirs were full of shredded pieces. By assuming the paperweight souvenir was full of shredded banknotes, the following calculations were performed using data from the fourth paperweight:

The weight of the shredded banknote pieces
= the weight of unopened paperweight – the weight of the empty cylinder
= (175.6 – 60) g
= 115.6 g

The number of equivalent banknotes
= the weight of the shredded banknotes pieces/the weight of one banknote
= 115.6/1.4
= 82.57 pieces

This cylinder only contained 82.57/138 = 60% of the shredded banknote pieces that the label had claimed. Although this issue is not the focus of this paper, it appears that the Hong Kong Monetary Authority has broken the law.

## VI. Limitations, improvements, and future development

Limitations:
Because the ground truth serial number was not the same as those of the shredded banknote pieces, manual placement of the pieces with the serial number was required. In addition, even though the shredded banknote pieces could construct a complete banknote, the serial number may not have come from the same banknote, and there is a high chance that it could not be exchanged for real money.

Improvements:
To increase the confidence in the location of a piece, another side of the piece could be used for analysis. If the locations of both sides were to correspond to the same position on the two sides of the banknote, it would produce much higher confidence in the location.

Future development:

In the current procedure, the rectangular template is created manually because the scanned piece may have defects such as a tear or crease. A machine learning program may be able to identify these defects and draw the template box.

An automated machine will be developed to pick up the shredded banknote pieces one by one and reconstruct the banknote.

## VII. CONCLUSION

Use of computer vision has demonstrated a high possibility of reconstructing banknotes from shredded banknote pieces. It was a surprise that the paperweight souvenir contained stones, and the number of shredded banknotes was much lower than 138 pieces of equivalent banknotes, as the label had claimed. The idea for this paper was discussed with the staff during my visit to the Hong Kong Monetary Authority visitor center. The paperweight souvenir is currently no longer available.

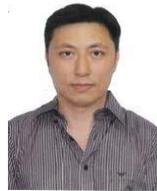

Chung To Kong is a PhD candidate at HKU and has worked in different industries, including cubeSat, robotics, LED screens, FPGAs, access control/CCTV, networking, and AI solutions. In 2017, he was a TEDx speaker about "Maker Mindset." He received his MSc in Electrical and Electronic Engineering from HKU and his bachelor's degree in Electrical Engineering and Computer Engineering with a minor in Commerce from the University of British Columbia in Vancouver, Canada.